\documentclass[journal,twoside,web]{ieeecolor}
\usepackage{jsen}
\usepackage{cite}
\usepackage{amsmath,amssymb,amsfonts}
\usepackage{algorithmic}
\usepackage{graphicx}
\usepackage{textcomp}
\usepackage{wrapfig}
\usepackage{verbatim}
\usepackage{subfig}
\usepackage{hyperref}
\newcommand{\pluseq}{\mathrel{+}\mathrel{\mkern-2mu}=}

\def\BibTeX{{\rm B\kern-.05em{\sc i\kern-.025em b}\kern-.08em
    T\kern-.1667em\lower.7ex\hbox{E}\kern-.125emX}}
\definecolor{abstractbg}{rgb}{0.89804,0.94510,0.83137}
\begin{document}
\title{Ant Colony Inspired Machine Learning Algorithm for Identifying and Emulating Virtual Sensors}
\author{Pranav Mani, E.S Gopi, \IEEEmembership{Senior Member, IEEE}, Koushik Kumaran, Hrishikesh Shekhar and Sharan Chandra
\thanks{(Pranav Mani, E.S Gopi, Koushik~Kumaran, Hrishikesh~Shekhar and Sharan~Chandra are co-first authors)}

\thanks{
Pranav Mani is with the National Institute of Technology, Tiruchirappalli, Tamil Nadu, 600031 INDIA (e-mail:pranavmani99@gmail.com). }
\thanks{Koushik Kumaran is with the National Institute of Technology, Tiruchirappalli, Tamil Nadu, 600031 INDIA (e-mail:koushikkumaran52@gmail.com).}
\thanks{Sharan Chandra is with the National Institute of Technology, Tiruchirappalli, Tamil Nadu, 600031 INDIA (e-mail:chandrasharan1@gmail.com).}
\thanks{Hrishikesh Shekhar is with the National Institute of Technology, Tiruchirappalli, Tamil Nadu, 600031 INDIA  (e-mail:hrishikeshshekhar@gmail.com).}
\thanks{Dr E.S Gopi is an Associate Professor with the department of Electronics and Communications Engineering at the National Institute of Technology, Tiruchirappalli, Tamil Nadu, 600031 INDIA
(e-mail:esgopi@nitt.edu).}}

\IEEEtitleabstractindextext{%
\fcolorbox{abstractbg}{abstractbg}{%
\begin{minipage}{\textwidth}%
\begin{wrapfigure}[17]{r}{5in}%
\includegraphics[width=5in]{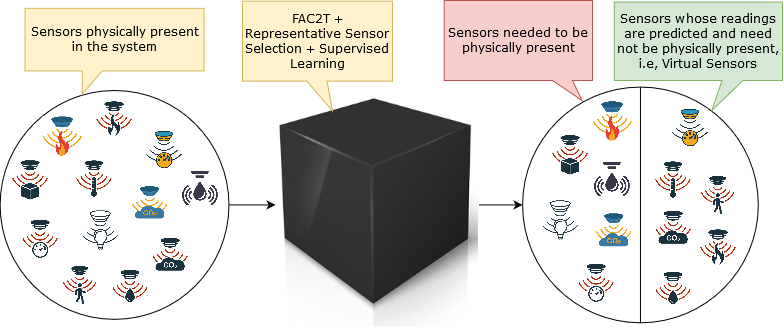}%
\end{wrapfigure}%
\begin{abstract}
The scale of systems employed in industrial environments demands a large number of sensors to facilitate meticulous monitoring and functioning. These requirements could potentially lead to inefficient system designs. The data coming from various sensors are often correlated due to the underlying relations in the system parameters that the sensors monitor. In theory, it should be possible to emulate the output of certain sensors based on other sensors. Tapping into such possibilities holds tremendous advantages in terms of reducing system design complexity. In order to identify the subset of sensors whose readings can be emulated, the sensors must be grouped into clusters. Complex systems generally have a large quantity of sensors that collect and store data over prolonged periods of time. This leads to the accumulation of massive amounts of data. In this paper we propose an end-to-end algorithmic solution, to realise virtual sensors in such systems. This algorithm splits the dataset into blocks and clusters each of them individually. It then fuses these clustering solutions to obtain a global solution using an Ant Colony inspired technique, FAC2T. Having grouped the sensors into clusters, we select representative sensors from each cluster. These sensors are retained in the system while the other sensors readings are emulated by applying supervised learning algorithms. 
\end{abstract}

\end{minipage}}}

\maketitle

\section{Introduction}\label{sec_introduction}
\IEEEPARstart{I}{ncreased} availability of computational power has encouraged analysis of system performance by examining large quantities of data. In most complex physical and cyber-physical systems \cite{baheti2011cyber}, relevant data is obtained from sensors embedded in the system design. To draw inferences from the collected data, most systems require a large number of sensors. However, in most systems, multiple sensor readings can be derived from each other due to the inherent relationships in the system parameters they monitor. 

Exploiting relationships between sensors allows for significant improvement in system design. In practice, it also leads to reduced power consumption and minimises maintenance and operational costs. In some critical systems, sensor failure can result in fatal mishaps \cite{FAABoeingGon}, \cite{ansaldi2016incidents}. Upon detecting sensor failure \cite{houck1998air}, \cite{kamal2014failure}, virtual sensors may then be used alongside physical sensors as a redundancy to account for the failed physical sensors.

Virtual sensors are not physically present while deploying the system in consideration. Instead, their readings are virtually derived from other sensors embedded in the system. This simpler system realisation requires identification of sensor clusters to understand the groups of sensors that are related to each other. It also requires knowledge of the mapping between observed sensor readings and the required virtual sensor readings.

In large-scale systems, the data collected over prolonged periods of time can be overwhelming to deal with. Applying a standard clustering algorithm on a dataset involving such high dimensional data is not feasible \cite{assent2012clustering}. Advanced computational capabilities are required for efficient and quick segregation of such high dimensional data into clusters. It is, therefore, necessary to design an algorithm for clustering, that is robust to such complications. 

There have been studies on proposing and using virtual sensors to serve several purposes in different operating environments. In \cite{NL1}, \cite{NL2}, the authors have explored model based methods, using non-linear state estimation techniques to develop virtual sensors for aircraft. However, these methods cannot be easily applied to other non-linear systems without the development of robust models for the particular system. In contrast, our proposed method does not require prior domain knowledge since it is a data driven approach. The use of neural networks to develop virtual sensors for pilot aid, diagnostics and emission monitoring was proposed in \cite{SPARK}. However, the authors assume prior knowledge of the target sensors whose values need to be emulated virtually. In our work, we identify the virtual sensors through clustering methods. In \cite{Entropy}, the authors use entropy based correlation to cluster sensors and identify representative sensors. However, they do not address the problem of large sampling rates leading to high dimensional data resulting in unstable cluster partitions. Further, the authors do not recover data of non-representative sensors from representative sensors through supervised learning techniques. In \cite{LSTMBidirectional}, the authors exploit the spatial and temporal relations between groups of sensors and make use of bidirectional LSTMs. However, the approach taken for selection of input sensors for prediction requires manual analysis of the correlations present in the system. In \cite{Sampling}, the authors make use of an adaptive sampling technique to address the problem of high sampling rates aimed at reducing the amount of energy consumed. Instead, by utilizing virtual sensors, we are able to obtain these sensor readings without sampling them.

The efficacy of using an Ant Colony based algorithm to maximise a given objective function for the task of clustering was demonstrated in \cite{Inspiration}. Inspired by this, we propose an algorithm, FAC2T(Fusion using Ant Colony Clustering Technique), to obtain an estimate of the globally optimal clustering solution by fusing the solutions obtained by applying the K-Means clustering algorithm\cite{bishop} on subsets of the data. Furthermore, FAC2T imposes a restriction on the number of clusters.

In this paper, we seek to establish an end-to-end algorithmic solution for emulating virtual sensors.
The major contributions of this paper are :
\begin{enumerate}
    \item FAC2T, An Ant Colony inspired algorithm to provide a heuristic method of clustering data belonging to extremely high-dimensional spaces.
    \item Providing an end-to-end algorithmic solution to emulate virtual sensors in complex systems
\end{enumerate}

\begin{figure}[!t]
\centering
    \includegraphics[width=0.5\textwidth]{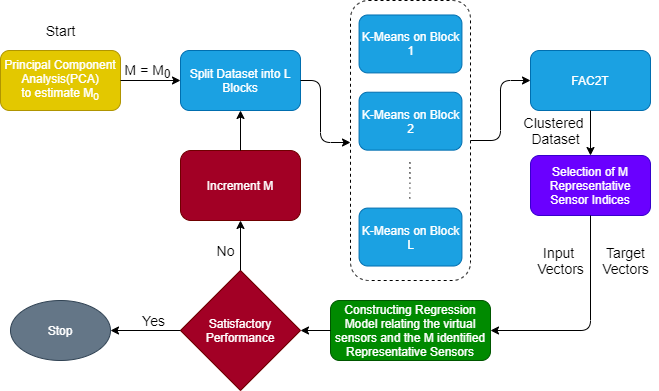}
    \caption{Proposed end-to-end algorithm. $M_0$ is the minimum number of required representative sensors obtained from PCA.}
    \label{fig_AlgoFlow}
\end{figure}

\subsection{Proposed Methodology} \label{sec_ProposedMethodology}
In this paper, we propose an end to end algorithm for tapping into the potential of virtual sensors. Section \ref{sec_PCA}, begins by presenting a rationale for the use of dimensionality reduction techniques to identify a starting point for the number of clusters the sensors need to be grouped into.

Section \ref{sec_Clustering} then presents a data fusion algorithm, FAC2T, for grouping the sensors into these clusters while avoiding the requirement for enormous computational power. We show that using FAC2T, initialized with clustering solutions obtained by applying K-Means algorithm on blocks of data, we can approach the global clustering solution.

In Section \ref{sec_ChoosingRepSensors}, we define a rule for the selection of a representative sensor from each of the clusters obtained using FAC2T. 

Subsequently, Section \ref{sec_Prediction} talks about employing supervised learning algorithms to predict the readings of virtual sensors using representative sensors. We pose the problem of mapping readings obtained from representative sensors to virtual sensors as a regression problem. We present 3 techniques: Linear Basis Function Regression(LBFR), Artificial Neural Networks(ANN) and Support Vector Regression(SVR) to tackle this supervised learning problem. A discussion is presented to briefly outline the factors to be considered in choosing a model based on the use case.
Finally, Section \ref{sec_results} present experiments and results of the algorithm along with their interpretations and scope for improvement.

The steps involved in our end-to-end algorithm are depicted in Fig. \ref{fig_AlgoFlow}.


\section{Principal Component Analysis to estimate minimum number of Representative Sensors}\label{sec_PCA}
We obtain an initialisation point for the number of clusters that the sensors should be grouped into using Principal Component Analysis (PCA) \cite{pca}. PCA helps us to identify M directions along which the K-dimensional vectors in the dataset have maximum variance. These directions are along the M eigenvectors corresponding to the M largest eigenvalues of the covariance matrix of the dataset. Using M as the number of clusters needed would be sufficient in the case where all the M-directions are along principal axes of the K-dimensional space. If this is not the case, we would need more than M principal axes vector to span the M dimensional space and hence more than M clusters. Therefore we use M as a starting point for the number of clusters and increment it until we achieve satisfactorily low estimation errors. 

\section{Fusion using Ant Colony Clustering Technique (FAC2T)}\label{sec_Clustering}

\subsection{Overview}
Given a dataset of considerably large size, it is not always feasible to perform cluster analysis on it. Hence, we split the dataset into L blocks and cluster each of them into M clusters using K-Means clustering. It is now required to fuse these individual solutions to obtain the best solution across the entire dataset.
Let $N_{sensors}$ be the total number of sensors. Now, given a set of L tentative clustering solutions, we use FAC2T to estimate the optimal solution for clustering $N_{sensors}$ into M clusters.

\subsection{FAC2T} \label{sec_ACOAdaptation}
 Ant Colony Optimization(ACO)\cite{aco} is a heuristic computational intelligence technique \cite{populationHeuristic}, \cite{gopipattern}. An ant colony is deployed to iteratively search and select optimal choices.
 
 Traditionally, ACO tries to emulate the behaviour of ant colonies by assigning "pheromone concentrations" to possible paths of travel. The ants then sample a path based on these "pheromone concentrations" in an attempt to reach their goal efficiently
 
 ACO has conventional applications in NP-hard combinatorial/optimization tasks such as the Travelling Salesman Problem \cite{NPHard}.
 
Given a set of L tentative solutions, we propose an algorithm based on the ant colony optimization to estimate the optimal solution of clustering N sensors into M clusters, based on a given objective function. We outline the algorithm below:   

\subsection{Definitions} \label{sec_definitions}
\subsubsection{Pheromone Table, $\textbf{P(n)}$}  \hfill\\
    The pheromone table at the $n^{th}$ iteration is a symmetric matrix of size $(N_{sensors} \times N_{sensors})$, represented as $\textbf{P(n)}$. Each element $P_{ij}(n)$ is an exponential moving average of metric values indicating the strength of a clustering in which the $i^{th}$ and $j^{th}$ sensor are clustered together.

\subsubsection{Representation of ants, $\mathbf{a_k(n)}$} \hfill\\
For each iteration, a total of $N_{ants}$ ants are maintained. Each ant is a clustering solution. The $j^{th}$ ant in the $n^{th}$ iteration is represented as $\mathbf{a_j(n)}$, a vector of size $N_{sensors}$. The $i^{th}$ element of $\mathbf{a_j(n)}$ represented as $a_{ji}(n)$ contains the index of the cluster to which the $i^{th}$ sensor belongs in the clustering solution $\mathbf{a_j(n)}$. $\mathbf{A(n)} \ = \left \{ \mathbf{a_j(n)}: \forall \ j \ \in \ \left\{1, 2, ..., N_{ants}\right \}  \right\}$.

\subsubsection{Representation of Clusters, $\mathbf{c_{kj}(n)}$} \hfill\\
The sensors in each ant $\mathbf{a_j(n)}$ are divided into M clusters. Each cluster $k$ in the ant $\mathbf{a_j(n)}$ is denoted by $\mathbf{c_{kj}(n)}$, which is a set consisting of the indices of the sensors belonging to the this cluster. The number of elements in $\mathbf{c_{kj}(n)}$ is given by the cardinality of the set, denoted by $|\mathbf{c_{kj}(n)}|$. 

\subsubsection{Hyper Parameters} \hfill\\
$\alpha$ - It corresponds to the rate of pheromone evaporation in the context of the Ant Colony Optimization. It is the fraction of the values in the pheromone matrix to be retained at the beginning of every iteration. 

$\beta$ - The number of sensors in each ant which are moved to other clusters based on a sampling of the pheromone matrix. These sensors are selected randomly and may vary across ants.

$\gamma$ - The number of iterations after which beta is decremented.

$\tau$ - Once in $\tau$ iterations, sensors are moved to randomly selected clusters (instead of using the pheromone matrix).

$\theta$ - The factor by which $\alpha$ is multiplied every iteration.

\subsubsection{Objective Function / Metric, $g(\mathbf{a_j(n)})$} \hfill\\
We define an objective function $g(\mathbf{a_j(n)})$ that evaluates an ant’s clustering. The aim of FAC2T is to obtain a clustering solution that maximises this Objective Function. The choice of the Objective Function depends on the user and the specific use case(for example, a correlation based Objective Function may be used to cluster correlated random variables). In this paper, we propose the following Objective Function:
\begin{equation}\label{eq_objfun}
    g(\mathbf{a_j(n)}) = \frac{1}{{\sum_{k=0}^{K}\mathbf{d(c_{kj}(n))}}}
\end{equation}
where $\mathbf{d(\mathbf{c_{kj}(n)})}$ evaluates the sum of squared distances between each sensor in cluster $\mathbf{c_{kj}(n)}$ and the centroid of $\mathbf{c_{kj}(n)}$.

Note that, in our proposed algorithm, the Objective Function for each ant is the sum of the Objective Function in \eqref{eq_objfun}, evaluated over all blocks in the data

\subsubsection{Adjacency Matrix} \hfill \\
The adjacency matrix $\mathbf{F(a_k(n))}$ is constructed by adding an edge for every pair of sensors that belong to the same cluster.
    
We define $\mathbf{F(a_k(n))}$, a symmetric matrix of size $(N_{sensors} \times N_{sensors})$:
\begin{equation*} \label{equation_1}
\mathbf{F(\mathbf{a_k(n)})_{ij}} = \left\{ \begin{array}{cc}
1 & \hspace{5mm} a_{ki}(n) = a_{kj}(n) \\
0 & \hspace{5mm} \text{otherwise} \\
\end{array} \right.
\end{equation*}
where $a_k(n)$ is the $k^{th}$ ant in the $n^{th}$ iteration and $i, j$ are the $i^{th}$ and $j^{th}$ sensors.
\\
The algorithmic procedure employed to model an Ant Colony in order to get the best clustering solution heuristically, is shown below:

 \begin{algorithmic}[1]
 \renewcommand{\algorithmicrequire}{\textbf{Inputs:}}
 \renewcommand{\algorithmicensure}{\textbf{Output:}}
  \REQUIRE \STATE {Data, K-Means clustering solutions on blocks of Data, Number of Clusters(M)}
 \ENSURE  Globally Optimal Clustering Solution \\
 \textit{\textbf{Initialization}}
    \STATE Each ant $\mathbf{a_j(0)}$ is initialized to the K-Means clustering solution of one block
   
    \STATE \textit{Pheromone Table} : $\mathbf{P(0)}$ is initialized with zeros
    \FOR {each ant $\mathbf{a_k(0)} \in \mathbf{A(0)}$}
         \STATE   $\mathbf{P(0)} \  \pluseq \ g(\mathbf{a_k(0)}) \times \mathbf{F(a_k(0))}$
    \ENDFOR
\\
 \textit{\textbf{Algorithm}}
 \FOR {n = 1,2....Iterations}
    \STATE $\mathbf{B(n)} = \mathbf{A(n)}$
    \FOR {each $\mathbf{a_k(n)} \in \mathbf{A(n)}$}
        \STATE Let $S$ be a set of random sensors, such that $|S|=\beta$ 
        \FOR {each sensor $s \in S$}
            \STATE 
                $P(X = j) = \left\{ \begin{array}{cc}
                \frac{1}{N_{sensors}} & \hspace{-4mm} n\mod{\tau}=0 \\ \\
                \frac{P_{sj}(n)}{\sum_{j=1}^{N_{sensors}} P_{sj}(n)} & \hspace{-4mm} \text{otherwise} \\
                \end{array} \right.
            $
            is sampled to get a sensor $j$.
            
            \STATE $x = a_{ks}(n)$
            \STATE $y = a_{kj}(n)$

            \IF {$\mathbf{|c_{xk}(n)|} = 1$}
                \STATE $z = \underset{z_1 \in \mathbf{c_{yk}(n)}}{\operatorname{argmin}} \sum\limits_{z_2 \in \mathbf{c_{yk}(n)}} P_{z_1z_2}(n)$
                \STATE $a_{kz}(n) = x$ 
            \ENDIF
            \STATE $a_{ks}(n) = y$. 
            
        \ENDFOR
    \ENDFOR
    
 \STATE $\mathbf{P(n + 1)} = \alpha \times \mathbf{P(n)}$
 \FOR {each $\mathbf{a_k(n)} \in \mathbf{A(n)}$}
     \STATE $ \mathbf{P(n + 1)} \pluseq (1-\alpha)\times g(\mathbf{a_k(n)})\times\mathbf{F(a_k(n))} $
 \ENDFOR
    \STATE $\mathbf{A(n+1)}$ = $N_{ants}$ ants from $\mathbf{A(n)}$ and $\mathbf{B(n)}$ with the highest metric values.
    
    \IF {$n\mod\gamma=0$}
        \STATE $\beta = \beta - 1$        
    \ENDIF
\STATE $\alpha = \alpha \times \theta $
\ENDFOR\\
 \end{algorithmic}
 
\begin{figure*}[!t]
\centering
    \includegraphics[width=0.78\textwidth]{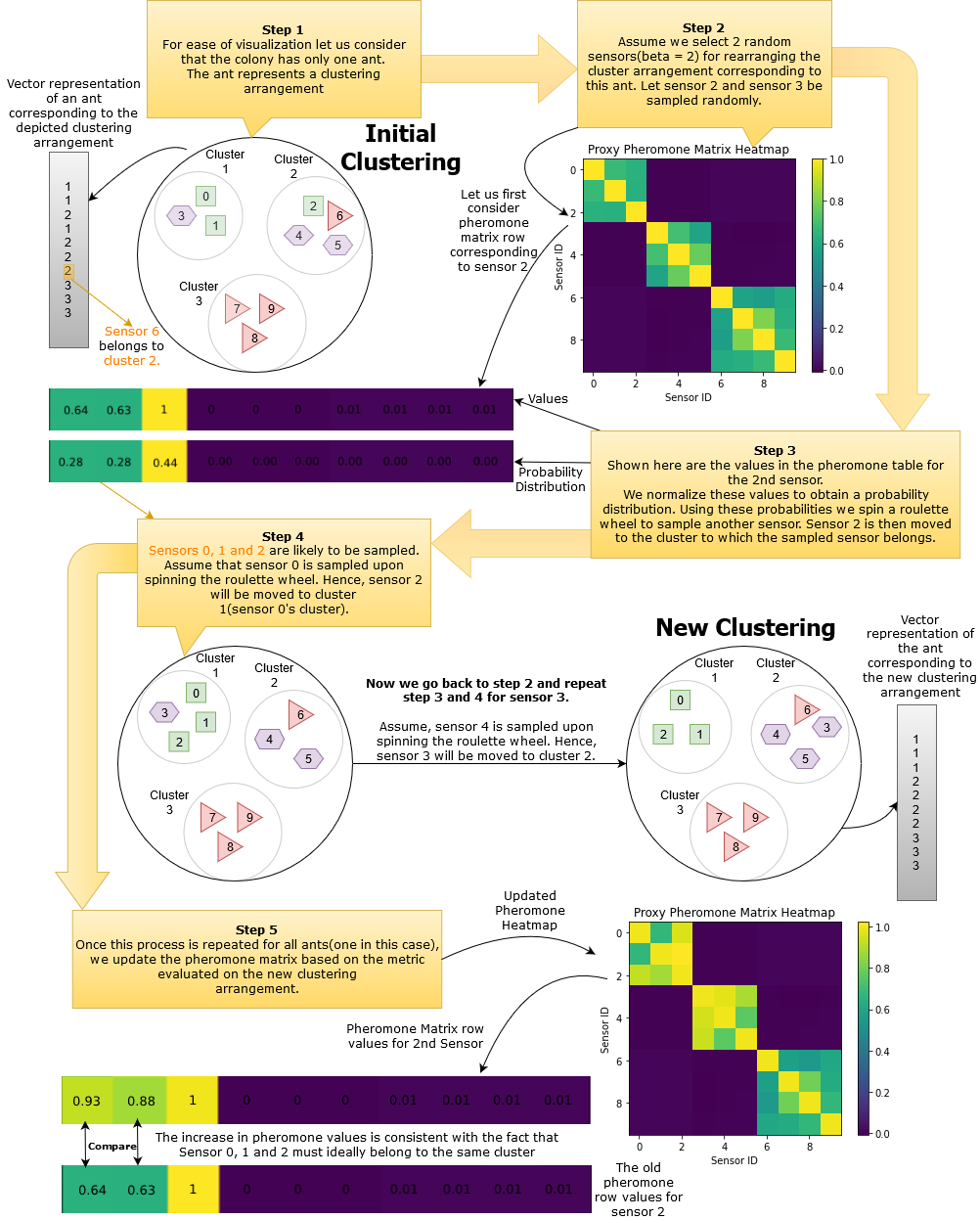}  
    \caption{The figure presents a visualization of the processes involved in one iteration performed by FAC2T, for one ant. We fix $\beta \ = \ 2$ and we have used a proxy pheromone matrix to impart the understanding. After the sampling and swapping process, the pheromone matrix scores are updated based on the evaluated metric. The algorithm, depicted above, is repeated for all ants in the colony(one in this case) to complete one iteration. Multiple iterations are performed until convergence is achieved.}
    \label{fig_ACOVis}
\end{figure*}

\subsection{Explanation of Algorithm}
Our algorithm takes as input K-Means clustering solutions on each block. We initialise ants with these K-Means clustering solutions according to the number of ants we want to use. In the case where the number of ants is greater than the number of blocks, there are $\lfloor N_{ants}/L \rfloor$ ants initialized to the clustering solution of each block, and the remaining ants initialized to solutions from random blocks. Otherwise, we evaluate the objective function on the clustering solution of each block and retain the best $N_{ants}$ solutions to be used as ant initializations. The pheromone table is initialized with zeros. 

Before we begin the search, we update the pheromone matrix with the metric evaluated on the initialized ants. Each iteration of the search for the best solution begins with creating a copy of the colony. We then mutate each ant by randomly selecting a subset of sensors whose clusters are to be swapped. For each candidate sensor belonging to this subset, we sample a pairing sensor using the pheromone values in the row of the pheromone matrix corresponding to the candidate sensor and move the candidate sensor to the pairing sensor's cluster. We note that, periodically, when the iteration becomes a multiple of $\tau$, this sampling of pairing sensor is done randomly.

An important point to note is that if the number of sensors in a cluster is 1 (a singleton cluster), then moving that sensor to another cluster will result in a reduction in the total number of clusters. In this case, the pairing sensor is chosen such that the cumulative pheromone scores of it with the other sensors in the cluster is minimum. Further, the clusters of the pairing and candidate sensors are swapped to maintain the total number of clusters.

We then evaluate the objective function on each mutated ant and selectively update the entries of the pheromone matrix. In order to achieve a stable trade-off between a history of explored solutions and a solution encountered in the present iteration, we retain a factor $\alpha$ of the values in the pheromone matrix and weigh down the newly evaluated metrics by a factor $1-\alpha$. This retaining of only a fraction of past values can be interpreted as an evaporation of pheromone values, and the disappearance of pheromones from sparsely explored solutions is crucial for the convergence of the algorithm.

After updating the pheromone matrix, we merge the copied colony with the mutated colony and retain $N_{ants}$ number of ants which correspond to the best metrics, to obtain the latest colony. We repeat these iterations until the metric saturates. We take this as an indication of convergence. We note that, periodically, we decrement the number of candidate sensors chosen to be swapped. This is driven by our expectation that as the quality of solutions increase, we need to swap lesser sensors in our search.

An intuitive visualization of the Ant Colony inspired Data Fusion based Clustering Algorithm can be seen in Fig. \ref{fig_ACOVis}.

\section{Choosing Representative Sensors} \label{sec_ChoosingRepSensors}
Using the clustering solution obtained from FAC2T, we choose a single sensor within each cluster to be a representative sensor. Only these representative sensors need to be retained physically and the other sensors are simply treated as virtual sensors. These representative sensors are included as input sensors to the supervised learning models for predicting the readings of the virtual sensors. In the $k^{th}$ cluster, let sensor vector $i$ be denoted as $\mathbf{s_{ki}}$. Let $n$ be the total number of sensors in this cluster. We evaluate the quality of each sensor as $Q(\mathbf{s_{ki}})$ which is defined as follows :

\begin{equation} \label{eq_quality}
    Q(\mathbf{s_{ki}}) =  \frac{1 - Var(\mathbf{c(\mathbf{s_{ki}})})}{1 - Mean(\mathbf{c(\mathbf{s_{ki}})})} \\
\end{equation}

where $\mathbf{c(\mathbf{s_{ki}})}$ is defined as follows :
\begin{equation*}
    \mathbf{c(\mathbf{s_{ki}})} =
        \begin{pmatrix}
            r(\mathbf{s_{ki}}, \mathbf{s_{k1}})\
            r(\mathbf{s_{ki}}, \mathbf{s_{k2}})\
            .\
            .\
            r(\mathbf{s_{ki}}, \mathbf{s_{kn}})\
        \end{pmatrix}^T
\end{equation*}   

$r(\mathbf{a}, \mathbf{b})$ used above is the Karl-Pearson Correlation Coefficient \cite{pearson1895correlation} between $\mathbf{a}$ and $\mathbf{b}$.

To choose a representative sensor for a cluster we calculate the "quality" of each sensor in a cluster and pick the sensor with the maximum quality within each cluster as that cluster's representative sensor.
    
The quality function is defined as above so that the chosen sensor will have a large average cross-correlation within the cluster, while the variance of these cross-correlations will be minimal.

\section{Predicting the readings of virtual sensors using representative sensors} \label{sec_Prediction}
In this section, we outline the supervised learning algorithms \cite{Liu2012} used to learn the mapping from observed readings of representative sensors to required readings of virtual sensors. Multiple techniques can be used to achieve this. For the purpose of our research, we have studied the performance of three machine learning techniques: 
\begin{enumerate}
    \item Linear Basis Function Regression (LBFR): 
    In LBFR\cite{montgomery2021introduction}, an input vector’s component values are a single reading from each of the representative sensors. We transform this vector to a higher dimensional feature space through the application of basis functions, to obtain feature vectors. In our study, we use the constant basis function, radial basis function and linear basis function. Our feature space is, therefore, of dimension M+2.
    \item Artificial Neural Networks (ANN):
    ANNs present a deep learning\cite{ann} approach to learn the required mapping. We use a standard fully connected neural network with 10 hidden layers and 50 neurons in all the hidden layers with ReLU activations\cite{relu}. We use the mean square error cost function across the predicted virtual sensor readings. For initializing the weights, we have used Xavier normal initialization \cite{xavier}, and they are updated using Adam optimiser\cite{adam}, with an initial learning rate of 0.001.
    \item Support Vector Regression (SVR):
    SVR \cite{awad2015support} is a flexible regression model that accommodates for a maximum acceptable error margin$(\epsilon)$. Furthermore, we specify a degree of deviation $C=1$ and $\epsilon = 0.1$, that controls the extent to which margins are adjusted to accommodate for deviations of data points from the margins.
\end{enumerate}

We note that LBFR allows much lesser model complexity than ANNs and SVR, and this translates to substantially lower training time. However, while we used a closed-form solution for obtaining the weights in our study, in other use cases where the feature vector space is of higher dimensions this may be computationally expensive. We would, in such cases, have to resort to gradient based methods which would increase training time. However, one of the main drawbacks of LBFR lies in choosing domain-specific basis functions to steer performance, whereas more sophisticated models such as ANNs can approximate these functions in a piecewise sense. However ANNs are trained using gradient based sequential learning methods, and require the tuning of various hyperparameters for smooth convergence. The selection of SVR is optimal when the use case necessitates an upper bound on the error. Any error above this is equally unaffordable in the system. SVR also allows a natural extension to the automation of the identification of this acceptable error threshold through other algorithms such as grid search or domain-specific techniques.

\section{Experiments and Results} \label{sec_results}
In this section, we showcase various performance metrics of our end-to-end algorithm, supplemented with relevant visualizations. In order to verify and demonstrate the robustness of FAC2T, we evaluate it on 100 synthetic datasets. We use synthetic datasets as we require an evaluation for FAC2T which is agnostic of the performance of regression models. Also, based on the data generation method used, we have an idea of the best possible clustering. We are therefore able to ensure consistent performance by observing performance across 100 synthetic datasets. We also run the entire algorithm on a pump dataset and aviation dataset, to evaluate the performance of the entire algorithm on real-world data.

\subsection{FAC2T Performance Validation} \label{sec_ACOVal}
\subsubsection{Synthetic Data Generation}\label{sec_SyntheticDataGen}
In order to validate and ensure of robustness of FAC2T, we test it on 100 synthetic datasets. We now outline how each synthetic dataset is generated.\\
We need clusters of sensor data vectors such that the vectors within-cluster are spaced closer to each other than vectors in other clusters. It is well known that the correlation between two vectors increases, the Euclidean distance between them decreases. Therefore, we generate sensor vectors which are highly correlated with other sensor vectors belonging to the same cluster.
This is done by using the Cholesky decomposition \cite{cholesky}. For each synthetic dataset, we generate a correlation matrix by first randomly picking sensor clusters. Having sampled sensor clusters, correlations between pairs of sensors in these clusters are obtained by sampling from a probability distribution that is designed to peak at a desired value and attenuate linearly in either direction from this point. For our analysis, we used a distribution that peaks at 0.75 and linearly falls to zero at 0.5 and 1 on either side. The correlation between any pair of sensors from two different clusters is set to zero. We now use the Cholesky decomposition \cite{cholesky} to obtain synthetic data. We would like to point out that before carrying out the Cholesky decomposition, it is necessary to perform the eigendecomposition of the correlation matrix and ensure its eigenvalues are greater than zero. For our analysis, we set a threshold of $1\mathrm{e}{-7}$ and any eigenvalue less than this is made equal to this threshold. We now reconstruct a positive definite correlation matrix by reconstructing the correlation matrix using the new eigenvalues.

An important point to note is that since the correlation matrix is created based upon the required grouping of sensors, we have access to the ideal clustering solution on the synthetic datasets.

\begin{figure*}[!t]
\centering
    \includegraphics[width=0.95\textwidth]{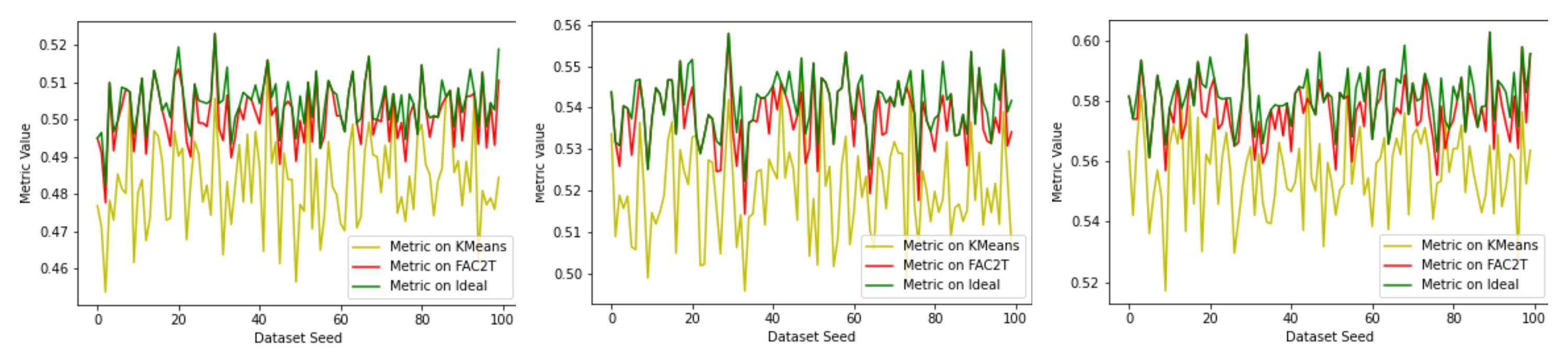}  
    \caption{Plot of the Objective Function, $g$, evaluated across 100 synthetic datasets for 20, 25 and 30 clusters respectively from left to right. In each plot, a comparison is drawn between Ideal Clustering, FAC2T Clustering and K-Means Clustering.}
    \label{fig_MetricPlotSynthetic}
\end{figure*}

\begin{figure}[!t]
\centering
\subfloat{\includegraphics[width=0.3\textwidth]{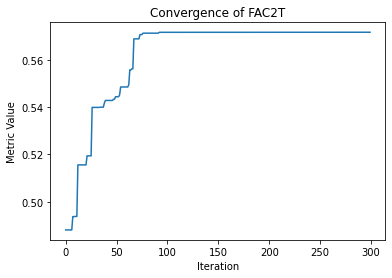}
\label{fig_ACOConv}}
\caption{Metric maximization performance of FAC2T}
\label{fig_HeatmapSynthetic}
\end{figure} 

\subsubsection{FAC2T Evaluation}\label{sec_ACOEval}
We use synthetic data to run two experiments to demonstrate the robustness of FAC2T

\textbf{Experiment A}

We use the method mentioned in Section \ref{sec_SyntheticDataGen} to generate 100 synthetic datasets. Each dataset consists of 100 sensors, with 70000 readings for each sensor. We split each of these datasets to form 70 blocks, each containing 1000 readings of every sensor
FAC2T is allowed to run for 200 iterations, with the hyperparameters set as follows:\\
$\alpha = 0.8$,
$\beta = 20$,
$\gamma = 20$,
$\delta = 15$,
$\tau = 10$,
$\theta = 1.007$

On completion of the algorithm, we select the ant(clustering solution) with the highest metric and compare it with the metric evaluated on the ideal clustering solution and the K-Means clustering solution on the entire dataset.

To visualize the consistency of the performance of FAC2T, the discussed procedure is repeated by varying the number of clusters as 20, 25, 30.

It can be seen from Fig. \ref{fig_MetricPlotSynthetic} that FAC2T consistently outperforms K-Means clusterings on the dataset. On observing closely, we can see that the metric performance on the clustering solution from FAC2T often matches that of the ideal clustering and as a result, the two can be seen to coincide at several points along the graph. 


\begin{figure*}[!t]
\centering
    \includegraphics[width=0.95\textwidth]{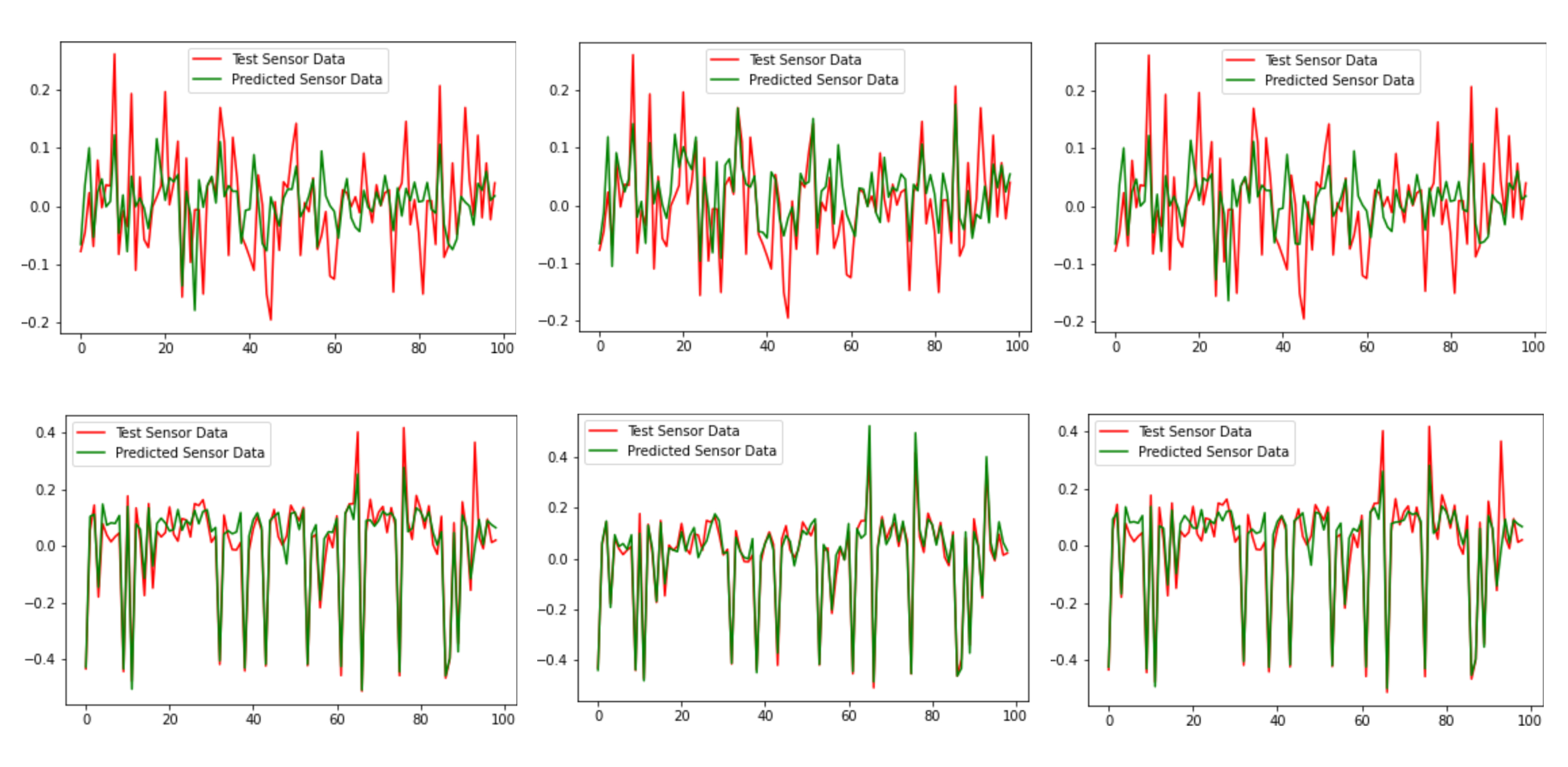}  
    \caption{Plots of Actual Sensor Data and Predicted Sensor Data on the Test Data for Sensor 01 and Sensor 33 are shown in the first and second row respectively. The plots from left to right represent LBFR, ANN and SVR respectively}
    \label{fig_PumpPredictions}
\end{figure*}

\begin{figure*}[!t]
\centering
    \includegraphics[width=0.96\textwidth]{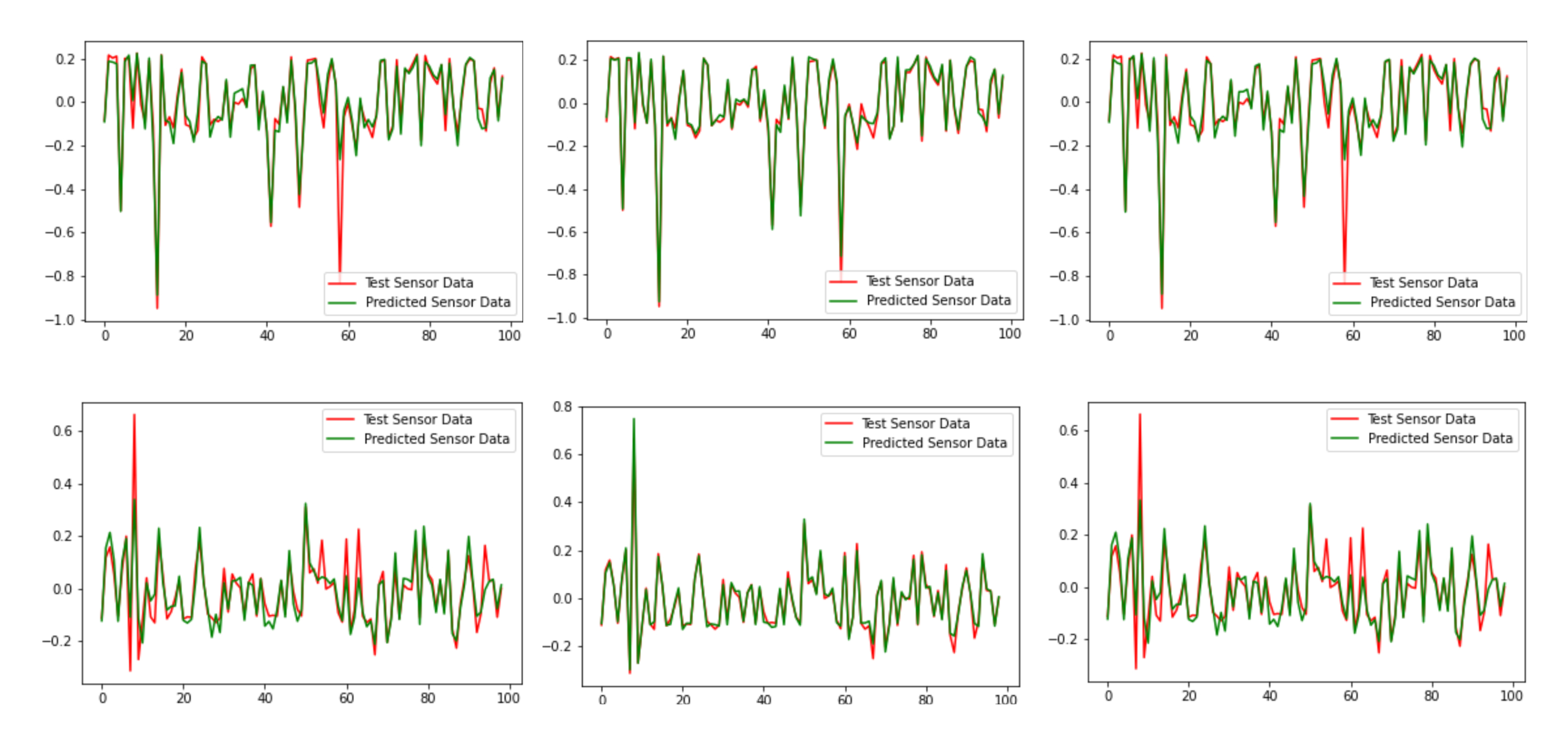}  
    \caption{Plots of Actual Sensor Data and Predicted Sensor Data on the Test Data for Exhaust Gas Temperature sensor and Pitch Angle LSP sensor are shown in the first and second row respectively. The plots from left to right represent LBFR, ANN and SVR respectively}
    \label{fig_AviationPredictions}
\end{figure*}

\textbf{Experiment B}

We use the method mentioned in Section \ref{sec_SyntheticDataGen} to generate 1 synthetic dataset consisting of 100 sensors, with 70000 readings for each sensor. The correlation matrix specified to create this dataset has 30 clusters of correlated sensors. We split this dataset to form 70 blocks, each containing 1000 readings of every sensor. We then run K-Means on each of these blocks. In the resulting K-Means clustering solutions, we randomly shuffle sensors between clusters. These corrupted clustering solutions are used to initialise the ants. We corrupt the initialisations to showcase the ability of FAC2T to correct even noisy initialisations. We then run FAC2T with the same hyperparameters as in Experiment A.

Fig. \ref{fig_HeatmapSynthetic} shows a plot of how the metric of the best ant varies with iterations when FAC2T is run on this synthetic dataset. From this graph, we can see that FAC2T succeeds in increasing the metric, and it saturates when the best ant represents the ideal clustering closely.

\subsection{Results on Real Datasets}\label{sec_ResultsOfPredictiveModels}
For validating our end-to-end algorithmic solution, we used two datasets:
\begin{itemize}
    \item Pump Dataset:
The dataset consists of 52 sensors concerning various operating parameters of a water pump. There are a total of around $2,00,000$ readings for each sensor. After the initial processing stage where null/missing values were removed, we obtained roughly $1,20,000$ data points. The dataset can be found at \cite{pump}. 
    \item Aviation Dataset:
The dataset consists of aircraft dynamics, system performance, and other engineering parameters governing the operation of a jet. After initial processing, we narrowed down 99 sensors for the purposes of our analysis. There is a maximum of $50,00,000$ readings for a few sensors, but many readings correspond to null/missing values in other sensor data. We, therefore, use $7,00,000$ data points where the corresponding data point for each sensor is obtained at the same time. The dataset can be found at \cite{DASH}.
\end{itemize}

\subsubsection{PCA results}
Using PCA, we obtain:
\begin{itemize}
    \item $M = 4$ for the Pump Dataset
    \item $M = 10$ for the Aviation Dataset 
\end{itemize}. We initialize our starting point as M clusters and proceed to obtain K-Means clustering solutions on blocks of our data, followed by an implementation of FAC2T.

\subsubsection{Model Performance} \label{sec_ModelPerformance}
In our analysis, FAC2T is used to cluster data from the datasets mentioned in Section \ref{sec_ResultsOfPredictiveModels}. We first cluster the sensors into clusters using FAC2T technique with K-means initialisation. Representative sensors are then selected from each cluster using the Quality function defined in Section \ref{sec_ChoosingRepSensors} by \eqref{eq_quality}. Subsequently, we train on various supervised learning models to learn this mapping from representative sensors to virtual sensors. We outline the results obtained for three choices of number of clusters: a) 7, 12, 17 b) 20, 25 and 30  for the pump dataset and the aviation dataset respectively. In order to visualize the performance, we also depict graphs showing readings for sensors predicted by our models versus actual readings, for a few randomly picked sensors. Fig. \ref{fig_PumpPredictions} is obtained from the pump dataset by fixing number of clusters as 12 while Fig. \ref{fig_AviationPredictions} is obtained from the aviation dataset by fixing number of clusters as 25.

We use mean square error on the testing data as measures of performance. These values are presented in Tables \ref{table_Mean1ErrorPump}, \ref{table_Mean1ErrorAviation}. For all the models we have used 80\% of the data for training and 20\% for testing.
\begin{table}[ht]
\centering
\begin{tabular}{|c|c|c|c|}
\hline
Model & 7 Clusters & 12 Clusters & 17 Clusters\\
\hline
\hline
LBFR & 0.0095527  & 0.0056825 & 0.0034265\\
ANN & 0.0053111 & 0.0030742 & 0.0017912\\
SVR & 0.0098801 & 0.0059130 & 0.0035294\\
\hline
\end{tabular}
\caption{Table showing Mean Square Error on Pump Dataset for predictive models on three different cluster numbers, LBFR: Linear basis function regression, ANN: Artificial neural networks, SVR: Support vector regression}
\label{table_Mean1ErrorPump}
\end{table}
\begin{table}[ht]
\centering
\begin{tabular}{|c|c|c|c|}
\hline
Model & 20 Clusters & 25 Clusters & 30 Clusters\\
\hline
\hline
LBFR & 0.0082150  & 0.0068895 & 0.0061271\\
ANN & 0.0050211 & 0.0042853 & 0.0040253\\
SVR & 0.0082311 & 0.0069098 & 0.0061470\\
\hline
\end{tabular}
\caption{Table showing Mean Square Error on Aviation Dataset for predictive models on three different cluster numbers, LBFR: Linear basis function regression, ANN: Artificial neural networks, SVR: Support vector regression}
\label{table_Mean1ErrorAviation}
\end{table}

From Table \ref{table_Mean1ErrorPump} and \ref{table_Mean1ErrorAviation}, we can see that, in line with our expectations, performance based on MSE improves with increasing number of clusters. We can also see that an appreciably low MSE, in order of 0.005, is achievable using ANN after removing nearly 87\% and 80\% of sensors from the systems respectively.
Further, the ANN has performed best albeit LBFR requires significantly lower training time.
\section{Conclusion and Future Scope}\label{sec_conclusion}
 In this paper, we demonstrate that it is possible to learn mappings between different sensors’ readings. We have shown in our analysis that a greatly reduced number of sensors still allows for obtaining all the sensors’ readings with a very marginal error, given that the inputs to the predictive models are selected so as to exploit this relationship. We verified the ability of FAC2T to obtain a clustering solution on the entire dataset while using K-Means clustering only on individual blocks of data. This allows for circumvention of the high computational requirements needed for clustering the entire dataset using standard clustering algorithms, while not compromising on the quality of the resultant clusters. Further, we have presented a study on various regression models that can be employed to solve the prediction problem.
 We note that further techniques may be explored to exploit the temporal
 relationships between and among data points

\bibliographystyle{IEEEtran}
\bibliography{VirtualSensREVISED}

\end{document}